\documentclass{article}
\usepackage{iclr2026_conference,times}

\usepackage{amsmath,amsfonts,bm}

\def\eqref#1{equation~\ref{#1}}

\def\1{\bm{1}}

\DeclareMathAlphabet{\mathsfit}{\encodingdefault}{\sfdefault}{m}{sl}
\SetMathAlphabet{\mathsfit}{bold}{\encodingdefault}{\sfdefault}{bx}{n}

\usepackage{graphicx}
\usepackage{booktabs}
\usepackage{float}
\usepackage{hyperref}
\usepackage{url}
\hypersetup{
  colorlinks=true,
  linkcolor=blue,
  citecolor=blue,
  urlcolor=blue
}

\iclrfinalcopy
\title{Where Bits Matter in World-Model Planning:\\A Paired Mixed-Bit Study for Efficient Spatial Reasoning}
\author{
Suraj Ranganath$^{*}$ \\
University of California San Diego \\
\texttt{suranganath@ucsd.edu}
\And
Anish Patnaik \\
University of California San Diego \\
\texttt{anpatnaik@ucsd.edu}
\AND
Vaishak Menon \\
University of California San Diego \\
\texttt{vamenon@ucsd.edu}
}

\begin{document}
\fancypagestyle{titlewithcorr}{
  \fancyhf{}
  \lhead{ArXiv preprint}
  \lfoot{\footnotesize $^{*}$Corresponding author}
  \cfoot{\thepage}
}
\maketitle
\thispagestyle{titlewithcorr}
\lhead{ArXiv preprint}

\begin{abstract}
Efficient spatial reasoning requires world models that remain reliable under tight precision budgets.
We study whether low-bit planning behavior is determined mostly by total bitwidth or by \emph{where} bits are allocated across modules.
Using DINO-WM on the Wall planning task, we run a paired-goal mixed-bit evaluation across uniform, mixed, asymmetric, and layerwise variants under two planner budgets.
We observe a consistent three-regime pattern: 8/6-bit settings remain close to FP16, 3-bit settings collapse, and 4-bit settings are allocation-sensitive.
In that transition region, preserving encoder precision improves planning relative to uniform quantization, and near-size asymmetric variants show the same encoder-side direction.
In a later strict 22-cell replication with smaller per-cell episode count, the mixed-vs-uniform INT4 sign becomes budget-conditioned, which further highlights the sensitivity of this transition regime.
These findings motivate module-aware, budget-aware quantization policies as a broader research direction for efficient spatial reasoning.
Code and run artifacts are available at \url{https://github.com/suraj-ranganath/DINO-MBQuant}.
\end{abstract}

\section{Introduction}
World-model planning has become a strong paradigm for sample-efficient control and spatial reasoning, especially when latent dynamics are built on pretrained visual features \citep{hafner2023dreamerv3,oquab2023dinov2,zhou2024dinowm}.
Deployment constraints, however, force these models into tight memory and latency budgets where aggressive quantization can destabilize behavior.
Recent broad surveys and world-model-specific studies report that low-bit degradation is often non-uniform across modules \citep{liu2025lowbitmodelquantizationdeep,fu2026quantwm}.

We ask whether, near low-bit operating points, planning quality is driven more by average precision or by how bits are allocated between the encoder and predictor. To answer this, we evaluate DINO-WM on the Wall task using paired-goal tests across a set of mixed-bit variants. Our empirical contributions identify when quantization is safe, when it degrades planning, and which allocation choices preserve performance under tight budgets. Concretely, we report three findings: (1) a three-regime pattern in which 8/6 bits remain similar to FP16, 4 bits form a sensitive transition, and 3 bits collapse; (2) the mixed-vs-uniform INT4 direction holds across two planner budgets and difficulty slices; and (3) asymmetric and layerwise ablations implicate the encoder as a primary sensitivity locus.

\section{Related Work and Broader Research Direction}
Mixed-precision quantization has long argued that sensitivity is layer-dependent rather than uniform, with automated policies that optimize hardware-constrained objectives \citep{wang2018haq,dong2019hawqv2}.
Recent low-bit post-training quantization for foundation models further supports this module-sensitivity view: practical 4--8 bit deployment typically depends on how precision is allocated, which channels are protected, and how outliers are handled \citep{dettmers2022llmint8,frantar2022gptq,xiao2022smoothquant,lin2023awq}.
Our results are consistent with that broader pattern, but in a planning setting where the metric is task success instead of perplexity or top-1 accuracy.

In world models, most efficiency work still emphasizes architecture or training design \citep{ha2018worldmodels,hafner2018planet,hafner2019dreamer,hafner2023dreamerv3}, while quantization-specific evidence is only beginning to appear \citep{fu2026quantwm,liu2025lowbitmodelquantizationdeep}.
This module-allocation perspective is also seen in efficient 3D spatial models such as FlatFormer \citep{liu2023flatformer}.
The present study contributes a paired-evaluation template for this gap.
The key observation is not that one specific mixed setting always wins, but that a transition regime exists where placement of bits changes planning outcomes.
For example, Table~\ref{tab:sizefair} shows that near-size encoder-heavy variants (E6/P4, E8/P4) improve over uniform INT4, while uniform INT6 demonstrates that higher total precision can still dominate under equal or slightly larger budgets.

This suggests a broader research direction for efficient spatial reasoning: treat bit allocation as a planning-aware resource-allocation problem, not a post-hoc compression detail.
A scalable agenda is to optimize encoder/predictor precision jointly under explicit memory and latency constraints, then test whether the same transition structure holds across environments, checkpoints, and world-model families.
If the structure persists, mixed-bit policies could become a first-class design axis for world-model deployment, analogous to horizon and optimizer budget choices.

\section{Setup}
\paragraph{Task and metric.}
We use pretrained DINO-WM \citep{zhou2024dinowm} on Wall and keep the repository-native planning success metric from planning logs.
We do not redefine task success.
All variants use weight-only linear-layer quantization and identical checkpoints.

The Wall environment requires spatial, goal-conditioned planning with obstacle and structure constraints; the planning success metric therefore reflects correct spatial reasoning in the model's latent rollouts.

We treat accuracy as the repository's native \texttt{success} metric (kept unchanged). Efficiency is reported via model storage (MB) and average planning/runtime per episode; planner tokens/steps budget is held fixed per the evaluation protocol below.

\paragraph{Model and hardware context.}
The model under study is DINO-WM with a visual encoder and latent predictor as defined in the original repository \citep{zhou2024dinowm}.
All experiments in this paper were executed locally on one Apple M4 MacBook Pro with 48GB unified memory.
Runs use PyTorch with MPS when available and CPU fallback otherwise, under a single fixed software stack across all variants.

\paragraph{Quantization implementation details.}
We use post-training, weight-only quantization on \emph{nn.Linear} layers.
For each output channel $j$ and target bitwidth $b$, weights are quantized as
$s_j=\max\{|W_j|\}/(2^{b-1}-1)$ and
$q_j=\mathrm{clip}(\mathrm{round}(W_j/s_j),-(2^{b-1}-1),2^{b-1}-1)$,
with dequantization $\tilde{W}_j=s_j q_j$ at inference.
This is symmetric per-output-channel quantization with no activation quantization, no calibration pass, and no retraining.

\paragraph{Variants.}
We evaluate FP16, uniform INT$\{8,6,4,3\}$, mixed INT$\{8,6,4,3\}$ (encoder FP16, predictor quantized), asymmetric variants (E8/P4, E6/P4, E4/P8, E4/P6), and an INT4 layerwise encoder-retention sweep (0\%, 25\%, 50\%, 75\%, 100\% encoder kept FP16; predictor fixed INT4).

\paragraph{Evaluation protocol.}
We use paired-goal evaluation: goals are generated once per seed and reused across variants.
Budget bA uses $(\texttt{goal\_H}=9,\texttt{opt\_steps}=2,\texttt{max\_iter}=2)$ with 3 seeds.
Budget bB uses $(12,3,3)$ with 2 seeds.
Each run has 10 episodes, yielding 65 runs and 650 episodes total.
The paired unit is $(\emph{seed},\emph{episode\_id})$: 30 paired units in bA and 20 in bB for the core mixed-vs-uniform comparisons.
We estimate paired deltas with 4000-sample bootstrap confidence intervals at paired-unit granularity and report two-sided sign-test $p$ values on non-tied pairs.
We report success, model size, runtime, paired deltas, and log-derived mechanistic signals.
In addition, we ran a strict follow-up block (run: m3\_main2exp\_v3) with 22 completed cells and 66 total episodes (3 episodes per cell), used here as a synchronization check between the original mixed-bit study and a lower-budget replication.
Code and run artifacts are available at \url{https://github.com/suraj-ranganath/DINO-MBQuant}.

\section{Main Results}
\begin{figure}[H]
  \centering
  \includegraphics[width=0.98\linewidth]{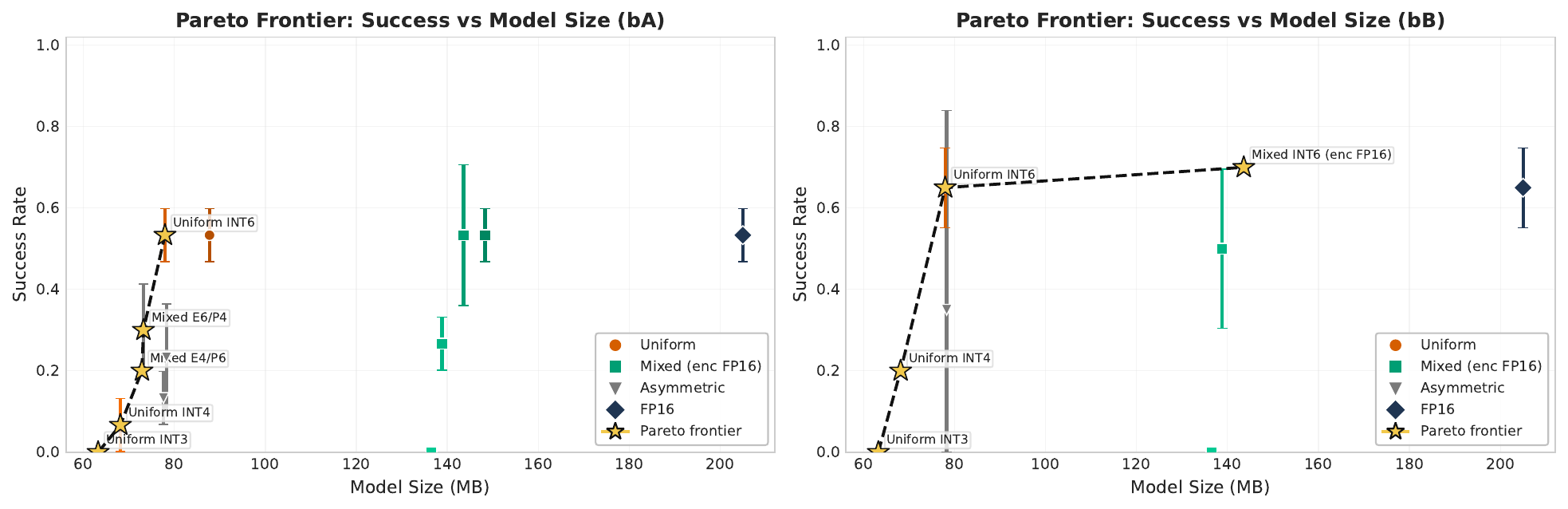}
  \caption{Success--size Pareto frontier for the paired mixed-bit study (budgets bA and bB). Each point is one variant, with vertical bars showing run-level 95\% confidence intervals over seeds. Stars denote non-dominated Pareto points (higher success, lower model size). The frontier shows a stable 8/6-bit region, an allocation-sensitive 4-bit transition, and a collapsed 3-bit region.}
  \label{fig:frontier}
\end{figure}

\paragraph{A three-regime pattern appears in this environment/checkpoint.}
Figure~\ref{fig:frontier} and Table~\ref{tab:main} show a stable regime at 8/6 bits, a transition at 4 bits, and collapse at 3 bits.
At budget bA, FP16, uniform INT8, mixed INT8, and uniform INT6 all reach 0.533 success.
At budget bB, FP16 and uniform INT6 both reach 0.650.
In contrast, all 3-bit variants are 0.0 at both budgets.

\paragraph{4-bit is allocation-sensitive.}
At 4 bits, mixed INT4 is higher than uniform INT4: 0.267 vs 0.067 at bA, and 0.500 vs 0.200 at bB.
Paired analysis gives +0.20 at bA (95\% CI [0.00, 0.40]) and +0.30 at bB ([0.00, 0.55]), with sign-test $p=0.109$ in both budgets.
We therefore interpret the 4-bit effect as \emph{directional} rather than statistically definitive (paired detail in Appendix Table~\ref{tab:paired} and forest view in Figure~\ref{fig:forest}).

\paragraph{Synchronous strict-run update.}
In the strict replication (22 cells, 66 episodes), the 8-bit-versus-4-bit regime split remains: FP16/mixed INT8/uniform INT8 are high, while 4-bit variants are lower overall.
However, mixed-vs-uniform INT4 is budget-conditioned in this smaller run: at bA, mixed INT4 exceeds uniform INT4 (0.333 vs 0.167; paired delta +0.167), while at bB, mixed INT4 is lower (0.000 vs 0.167; paired delta -0.167).
Episode-level matchup counts are balanced in pooled view (1 mixed-only win vs 1 uniform-only win; Appendix Table~\ref{tab:strictmatch}).
We interpret this as a sensitivity signal near the 4-bit frontier: allocation matters, but direction can flip with budget under low sample size.

\paragraph{Fairness check: allocation versus total precision budget.}
Mixed INT4 (138.84 MB) is substantially larger than uniform INT4 (68.12 MB), so mixed-vs-uniform alone does not isolate allocation from total precision.
To partially control this, we compare near-size asymmetric variants at bA: E6/P4 (73.19 MB, success 0.300) and E8/P4 (78.25 MB, 0.233) both improve over uniform INT4 (68.12 MB, 0.067), while remaining far smaller than mixed INT4.
This suggests encoder-side bits are useful even under tight size budgets, while higher total precision (for example uniform INT6 at 77.92 MB, success 0.533) remains another strong driver (size-aware context in Appendix Table~\ref{tab:sizefair} and map in Figure~\ref{fig:asym}).

\paragraph{Asymmetric and layerwise evidence supports encoder sensitivity.}
At bA, E6/P4 reaches 0.300 success at 73.19 MB, improving over uniform INT4 (0.067 at 68.12 MB), with paired delta +0.233 (95\% CI [0.033, 0.433]).
By contrast, increasing predictor bits while keeping encoder at INT4 (E4/P8, E4/P6) does not match mixed INT4.
Numerically, E4/P8 is 0.133 with paired delta $-0.133$ relative to mixed INT4, suggesting that spending extra bits on the predictor cannot compensate for encoder degradation at this frontier.
Layerwise INT4 ablation (Appendix Figure~\ref{fig:enc_curve}) is non-monotonic but peaks when the encoder is fully preserved (0.25 at 100\% FP16), supporting the hypothesis that encoder precision is often a key bottleneck near the 4-bit frontier.

\paragraph{Mechanistic diagnostics are consistent with representation degradation.}
Across 65 run-level points (variant $\times$ budget $\times$ seed), success correlates negatively with divergence metrics from planning logs: Spearman $\rho=-0.928$ for mean state distance and $\rho=-0.708$ for visual-embedding divergence (Appendix Figure~\ref{fig:mech}).
This does not prove causality, but it supports a plausible mechanism: low-bit encoder degradation weakens latent geometry used for goal-directed planning.

\begin{table}[H]
\centering
\small
\begin{tabular}{lccc}
\toprule
Variant & Success bA & Success bB & Size (MB) \\
\midrule
FP16 & 0.533 & 0.650 & 204.99 \\
Uniform INT6 & 0.533 & 0.650 & 77.92 \\
Mixed INT6 & 0.533 & 0.700 & 143.58 \\
Uniform INT4 & 0.067 & 0.200 & 68.12 \\
Mixed INT4 & \textbf{0.267} & \textbf{0.500} & 138.84 \\
Uniform INT3 & 0.000 & 0.000 & 63.23 \\
Mixed INT3 & 0.000 & 0.000 & 136.47 \\
\bottomrule
\end{tabular}
\caption{Core mixed-bit outcomes from the primary paired run. Means are shown; uncertainty and paired statistics are reported in Appendix Tables~\ref{tab:paired} and \ref{tab:sizefair}.}
\label{tab:main}
\end{table}

\section{Discussion, Limitations, and Outlook}
The data supports a narrower claim than ``mixed is always better'': near low-bit transition points, planning quality depends on both total precision and module allocation. In the primary paired run, mixed and encoder-heavy asymmetric variants outperform uniform INT4, while all INT3 variants collapse. This pattern suggests encoder precision is a key control variable in the 4-bit regime.

The asymmetric results raise a practical design question for follow-up work: can budget-constrained bit allocation be optimized directly for planning success rather than model reconstruction error? The non-monotonic layerwise curve suggests a second question: which encoder blocks are precision-critical, and does that set change across environments? The strong divergence-success correlations further motivate causal tests such as representation-preserving calibration and geometry-aware quantization objectives.

Generalization here is operational, not cross-domain. The primary mixed-bit run shows directional mixed-vs-uniform INT4 gains across two planner budgets and across goal-distance bins (Appendix Table~\ref{tab:difficultybins}), while the strict low-sample replication shows a budget-conditioned split (Appendix Table~\ref{tab:strictmatch}). Taken together, these runs suggest that the 4-bit boundary is highly sensitive to planning budget and sample regime, rather than governed by a single monotonic rule.

This is a preliminary study with clear constraints. We evaluate one environment/checkpoint family with modest paired sample sizes, so uncertainty near INT4 remains substantial and mixed-vs-uniform INT4 is still partially confounded by size. We also evaluate only weight-only PTQ (no activation quantization, calibration, or retraining). All runs were executed on one M4 MacBook Pro (48GB), which constrained sweep breadth.

Future work will run the same paired template on more environments and model families, with larger seed/episode counts and stricter size matching. We will test activation-aware PTQ, calibration, and QAT, and replicate on NVIDIA GPUs with hardware kernels to characterize deployment tradeoffs more fully. Beyond module-level analysis, we will run finer block-wise and attention-head localization with automated sensitivity ranking. The long-term goal is to optimize bit allocation directly for planning success under explicit memory and latency constraints.

\section{Conclusion}
In a paired mixed-bit study on DINO-WM planning, we find structured low-bit behavior: 8/6 bits are mostly safe, 3 bits collapse, and 4 bits is the sensitive transition regime.
At 4 bits, mixed and asymmetric results consistently point toward encoder-side precision being more valuable than predictor-side precision, although effect sizes remain statistically limited and partially confounded by total precision budget.
These findings motivate module-aware, budget-aware quantization policies as a concrete research direction for efficient spatial reasoning.

\bibliography{references}

@misc{zhou2024dinowm,
  doi = {10.48550/ARXIV.2411.04983},
  url = {https://arxiv.org/abs/2411.04983},
  author = {Zhou, Gaoyue and Pan, Hengkai and LeCun, Yann and Pinto, Lerrel},
  title = {DINO-WM: World Models on Pre-trained Visual Features enable Zero-shot Planning},
  publisher = {arXiv},
  year = {2024},
  copyright = {Creative Commons Attribution 4.0 International}
}

@misc{fu2026quantwm,
  doi = {10.48550/ARXIV.2602.02110},
  url = {https://arxiv.org/abs/2602.02110},
  author = {Fu, Zhongqian and Zhao, Tianyi and Han, Kai and Zhou, Hang and Chen, Xinghao and Wang, Yunhe},
  title = {An Empirical Study of World Model Quantization},
  publisher = {arXiv},
  year = {2026},
  copyright = {arXiv.org perpetual, non-exclusive license}
}

@misc{ha2018worldmodels,
  doi = {10.48550/ARXIV.1803.10122},
  url = {https://arxiv.org/abs/1803.10122},
  author = {Ha, David and Schmidhuber, J{\"u}rgen},
  title = {World Models},
  publisher = {arXiv},
  year = {2018},
  copyright = {Creative Commons Attribution 4.0 International}
}

@misc{hafner2018planet,
  doi = {10.48550/ARXIV.1811.04551},
  url = {https://arxiv.org/abs/1811.04551},
  author = {Hafner, Danijar and Lillicrap, Timothy and Fischer, Ian and Villegas, Ruben and Ha, David and Lee, Honglak and Davidson, James},
  title = {Learning Latent Dynamics for Planning from Pixels},
  publisher = {arXiv},
  year = {2018},
  copyright = {arXiv.org perpetual, non-exclusive license}
}

@misc{hafner2019dreamer,
  doi = {10.48550/ARXIV.1912.01603},
  url = {https://arxiv.org/abs/1912.01603},
  author = {Hafner, Danijar and Lillicrap, Timothy and Ba, Jimmy and Norouzi, Mohammad},
  title = {Dream to Control: Learning Behaviors by Latent Imagination},
  publisher = {arXiv},
  year = {2019},
  copyright = {arXiv.org perpetual, non-exclusive license}
}

@misc{hafner2023dreamerv3,
  doi = {10.48550/ARXIV.2301.04104},
  url = {https://arxiv.org/abs/2301.04104},
  author = {Hafner, Danijar and Pasukonis, Jurgis and Ba, Jimmy and Lillicrap, Timothy},
  title = {Mastering Diverse Domains through World Models},
  publisher = {arXiv},
  year = {2023},
  copyright = {Creative Commons Attribution 4.0 International}
}

@misc{oquab2023dinov2,
  doi = {10.48550/ARXIV.2304.07193},
  url = {https://arxiv.org/abs/2304.07193},
  author = {Oquab, Maxime and Darcet, Timoth{\'e}e and Moutakanni, Th{\'e}o and Vo, Huy and Szafraniec, Marc and Khalidov, Vasil and Fernandez, Pierre and Haziza, Daniel and Massa, Francisco and El-Nouby, Alaaeldin and Assran, Mahmoud and Ballas, Nicolas and Galuba, Wojciech and Howes, Russell and Huang, Po-Yao and Li, Shang-Wen and Misra, Ishan and Rabbat, Michael and Sharma, Vasu and Synnaeve, Gabriel and Xu, Hu and Jegou, Herv{\'e} and Mairal, Julien and Labatut, Patrick and Joulin, Armand and Bojanowski, Piotr},
  title = {DINOv2: Learning Robust Visual Features without Supervision},
  publisher = {arXiv},
  year = {2023},
  copyright = {arXiv.org perpetual, non-exclusive license}
}

@misc{dettmers2022llmint8,
  doi = {10.48550/ARXIV.2208.07339},
  url = {https://arxiv.org/abs/2208.07339},
  author = {Dettmers, Tim and Lewis, Mike and Belkada, Younes and Zettlemoyer, Luke},
  title = {LLM.int8(): 8-bit Matrix Multiplication for Transformers at Scale},
  publisher = {arXiv},
  year = {2022},
  copyright = {Creative Commons Attribution 4.0 International}
}

@misc{frantar2022gptq,
  doi = {10.48550/ARXIV.2210.17323},
  url = {https://arxiv.org/abs/2210.17323},
  author = {Frantar, Elias and Ashkboos, Saleh and Hoefler, Torsten and Alistarh, Dan},
  title = {GPTQ: Accurate Post-Training Quantization for Generative Pre-trained Transformers},
  publisher = {arXiv},
  year = {2022},
  copyright = {Creative Commons Attribution 4.0 International}
}

@misc{xiao2022smoothquant,
  doi = {10.48550/ARXIV.2211.10438},
  url = {https://arxiv.org/abs/2211.10438},
  author = {Xiao, Guangxuan and Lin, Ji and Seznec, Mickael and Wu, Hao and Demouth, Julien and Han, Song},
  title = {SmoothQuant: Accurate and Efficient Post-Training Quantization for Large Language Models},
  publisher = {arXiv},
  year = {2022},
  copyright = {Creative Commons Attribution 4.0 International}
}

@misc{lin2023awq,
  doi = {10.48550/ARXIV.2306.00978},
  url = {https://arxiv.org/abs/2306.00978},
  author = {Lin, Ji and Tang, Jiaming and Tang, Haotian and Yang, Shang and Chen, Wei-Ming and Wang, Wei-Chen and Xiao, Guangxuan and Dang, Xingyu and Gan, Chuang and Han, Song},
  title = {AWQ: Activation-aware Weight Quantization for LLM Compression and Acceleration},
  publisher = {arXiv},
  year = {2023},
  copyright = {arXiv.org perpetual, non-exclusive license}
}

@misc{liu2023flatformer,
  doi = {10.48550/ARXIV.2301.08739},
  url = {https://arxiv.org/abs/2301.08739},
  author = {Liu, Zhijian and Yang, Xinyu and Tang, Haotian and Yang, Shang and Han, Song},
  title = {FlatFormer: Flattened Window Attention for Efficient Point Cloud Transformer},
  publisher = {arXiv},
  year = {2023}
}

@misc{liu2025lowbitmodelquantizationdeep,
  title={Low-bit Model Quantization for Deep Neural Networks: A Survey},
  author={Liu, Kai and Zheng, Qian and Tao, Kaiwen and Li, Zhiteng and Qin, Haotong and Li, Wenbo and Guo, Yong and Liu, Xianglong and Kong, Linghe and Chen, Guihai and Zhang, Yulun and Yang, Xiaokang},
  year={2025},
  eprint={2505.05530},
  archivePrefix={arXiv},
  primaryClass={cs.LG},
  url={https://arxiv.org/abs/2505.05530}
}

@misc{wang2018haq,
  doi = {10.48550/ARXIV.1811.08886},
  url = {https://arxiv.org/abs/1811.08886},
  author = {Wang, Kuan and Liu, Zhijian and Lin, Yujun and Lin, Ji and Han, Song},
  title = {HAQ: Hardware-Aware Automated Quantization with Mixed Precision},
  publisher = {arXiv},
  year = {2018}
}

@misc{dong2019hawqv2,
  doi = {10.48550/ARXIV.1911.03852},
  url = {https://arxiv.org/abs/1911.03852},
  author = {Dong, Zhen and Yao, Zhewei and Gholami, Amir and Mahoney, Michael and Keutzer, Kurt},
  title = {HAWQ-V2: Hessian Aware Trace-Weighted Quantization of Neural Networks},
  publisher = {arXiv},
  year = {2019}
}
\bibliographystyle{iclr2026_conference}

\appendix
\clearpage
\section{Appendix: Extended Results with Context}

\subsection{How to Read the Appendix}
The main paper establishes the core claim from one frontier figure and one compact table.
This appendix provides supporting analyses that answer three reviewer-facing questions:
\textbf{(i)} Is the 4-bit effect robust to planner budget?
\textbf{(ii)} Does evidence localize precision sensitivity to the encoder?
\textbf{(iii)} Are observed gains consistent with a representation-degradation mechanism?

\subsection{Paired Mixed-Bit Effect Table}
Table~\ref{tab:paired} reports paired deltas for the most relevant comparisons.
The first two rows are the primary claim (mixed INT4 vs uniform INT4 at bA and bB).
The asymmetric rows test whether increasing encoder bits or predictor bits is more useful at similar model sizes.

\begin{table}[H]
\centering
\small
\begin{tabular}{lcccc}
\toprule
Comparison & $n$ pairs & Delta & 95\% CI & Sign-test $p$ \\
\midrule
bA: mixed\_int4 - uniform\_int4 & 30 & +0.200 & [0.000, 0.400] & 0.109 \\
bB: mixed\_int4 - uniform\_int4 & 20 & +0.300 & [0.000, 0.550] & 0.109 \\
bA: enc6\_pred4 - uniform\_int4 & 30 & +0.233 & [0.033, 0.433] & 0.065 \\
bA: enc4\_pred8 - mixed\_int4 & 30 & -0.133 & [-0.333, 0.067] & 0.344 \\
\bottomrule
\end{tabular}
\caption{Paired effect estimates from the consolidated mixed-bit pairwise statistics file.}
\label{tab:paired}
\end{table}

\noindent\textbf{Inference.}
Rows 1--2 show that mixed INT4 consistently outperforms uniform INT4 under paired episodes in both budgets, but with moderate uncertainty.
Rows 3--4 strengthen the allocation interpretation: encoder-heavy E6/P4 improves over uniform INT4, while predictor-heavy E4/P8 does not recover mixed INT4.
Together, the table supports an encoder-priority hypothesis near the 4-bit boundary.

\subsection{Strict Replication Matchup Context}
The strict run was designed as a low-cost synchronization check.
Table~\ref{tab:strictmatch} reports deduplicated episode-level matchup counts for mixed INT4 vs uniform INT4.

\begin{table}[H]
\centering
\small
\begin{tabular}{lccccc}
\toprule
Scope & Paired eps & Mixed wins & Uniform wins & Both win & Both fail \\
\midrule
bA & 6 & 1 & 0 & 1 & 4 \\
bB & 6 & 0 & 1 & 0 & 5 \\
Pooled (bA+bB) & 12 & 1 & 1 & 1 & 9 \\
\bottomrule
\end{tabular}
\caption{Strict replication matchup contingency; pooled mixed-vs-uniform win differential is zero.}
\label{tab:strictmatch}
\end{table}

\noindent\textbf{Inference.}
This strict replication preserves the broader regime structure (8-bit stronger than 4-bit) but shows that mixed-vs-uniform INT4 direction can depend on budget at small sample size.
This is consistent with treating the 4-bit zone as a sensitive transition regime rather than a single fixed ordering.

\subsection{Size-Fairness and Allocation Context}
Table~\ref{tab:sizefair} makes the size confound explicit.
Mixed INT4 is much larger than uniform INT4, so we also compare near-size asymmetric variants.
At bA, E6/P4 and E8/P4 are close to uniform INT4 in size and still improve success, indicating that encoder-side allocation helps even without fully preserving the encoder in FP16.

\begin{table}[H]
\centering
\small
\begin{tabular}{lccc}
\toprule
Variant (bA unless noted) & Success & Size (MB) & Note \\
\midrule
Uniform INT4 & 0.067 & 68.12 & baseline low-size 4-bit \\
E6/P4 & 0.300 & 73.19 & near-size encoder-heavy \\
E8/P4 & 0.233 & 78.25 & near-size encoder-heavy \\
Uniform INT6 & 0.533 & 77.92 & higher total precision \\
Mixed INT4 & 0.267 & 138.84 & larger precision budget \\
\midrule
Uniform INT4 (bB) & 0.200 & 68.12 & budget-shift check \\
E8/P4 (bB) & 0.350 & 78.25 & near-size improvement \\
Mixed INT4 (bB) & 0.500 & 138.84 & larger precision budget \\
\bottomrule
\end{tabular}
\caption{Size-aware view of allocation effects. Mixed INT4 is stronger than uniform INT4 but uses substantially more total precision. Near-size asymmetric variants still show encoder-side gains over uniform INT4.}
\label{tab:sizefair}
\end{table}

\noindent\textbf{Inference.}
This table is the key fairness check for Figure~\ref{fig:frontier}.
If the mixed INT4 gain were only due to larger size, near-size asymmetric variants would be expected to match uniform INT4.
Instead, encoder-heavy near-size variants improve over uniform INT4, indicating that \emph{where} bits are allocated matters in addition to total precision.

\subsection{Budget Robustness Context}
\begin{figure}[H]
  \centering
  \includegraphics[width=0.95\linewidth]{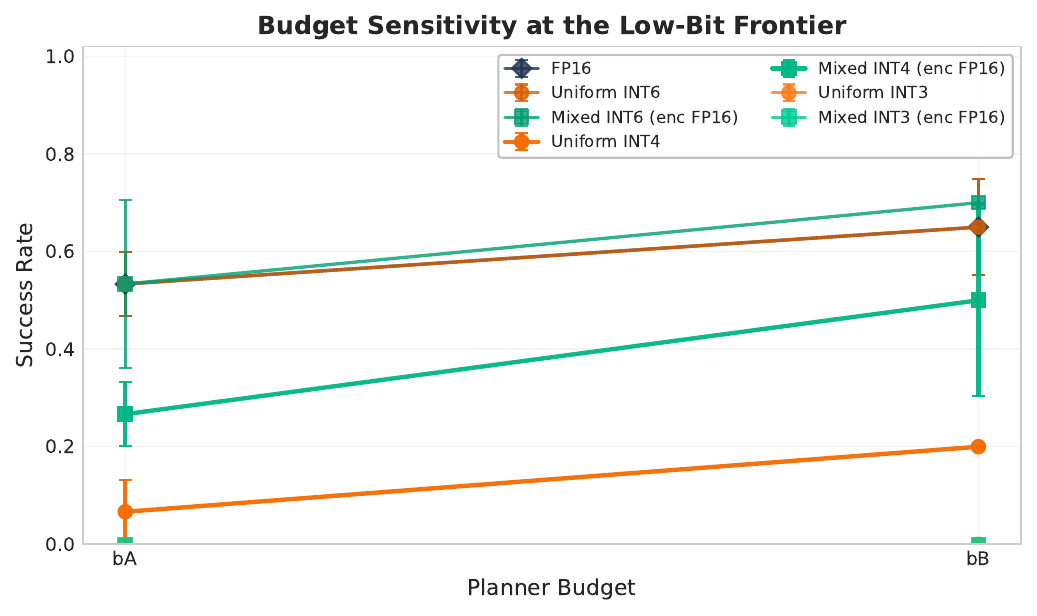}
  \caption{Budget robustness view. Uniform INT3 remains collapsed; mixed INT4 remains above uniform INT4 at both budgets.}
  \label{fig:budget}
\end{figure}

Figure~\ref{fig:budget} shows that the mixed-vs-uniform INT4 gap persists when planner budget increases from bA to bB.
This reduces the likelihood that the observed gap is only an artifact of one optimization budget.

\noindent\textbf{Inference.}
The mixed-over-uniform INT4 direction survives when the planner budget increases from bA to bB.
This reduces the chance that the headline finding is a single-budget artifact and supports treating the 4-bit transition as a structural regime.

\subsection{Encoder-Localization Context}
\begin{figure}[H]
  \centering
  \includegraphics[width=0.95\linewidth]{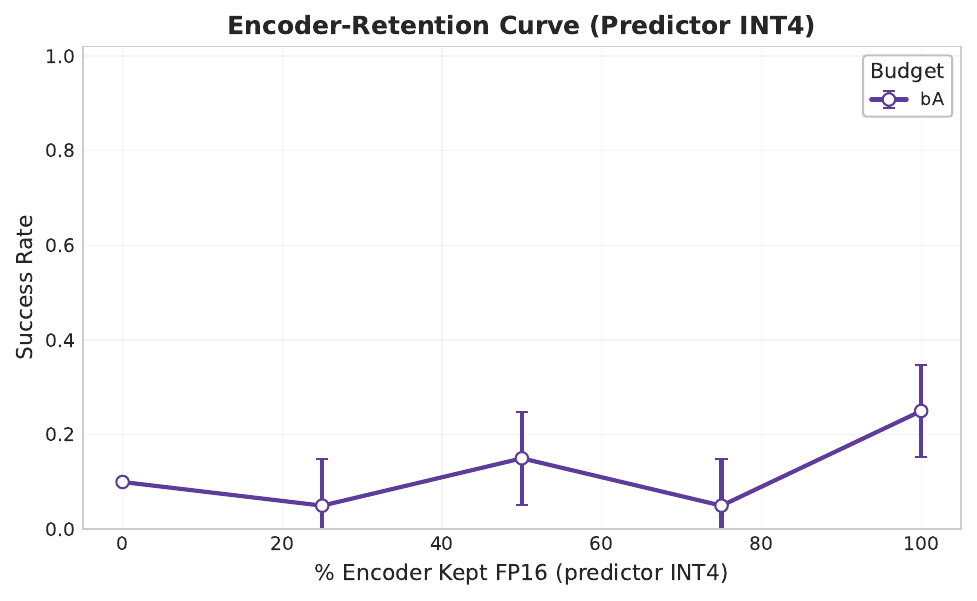}
  \caption{Encoder-retention sweep (INT4 predictor). The highest mean success occurs when encoder precision is fully preserved.}
  \label{fig:enc_curve}
\end{figure}

Figure~\ref{fig:enc_curve} gives a coarse localization test: with predictor fixed at INT4, preserving more encoder precision improves peak success, although intermediate points are non-monotonic.
This supports the main claim while motivating finer block-level studies.

\noindent\textbf{Inference.}
Even in a coarse 5-point sweep, success rises toward the fully preserved encoder endpoint.
Although non-monotonicity remains, the curve indicates that failures at 4-bit are not uniformly distributed across modules and that encoder precision is a dominant control variable.

\subsection{Difficulty-Conditioned Context}
\begin{figure}[H]
  \centering
  \includegraphics[width=0.95\linewidth]{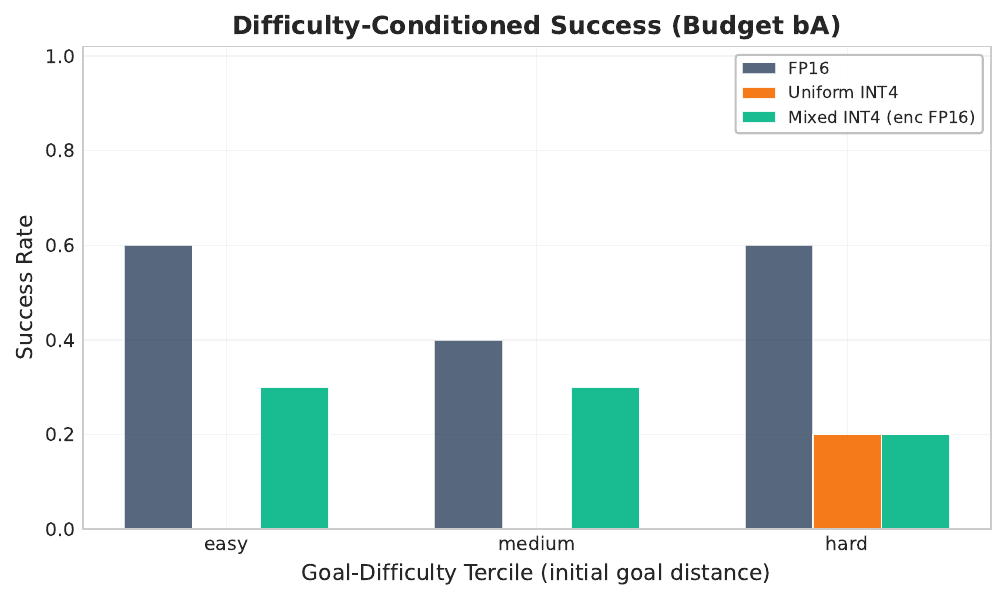}
  \caption{Difficulty-conditioned success at bA (paired episodes). Mixed INT4 is higher in most plotted bins.}
  \label{fig:difficulty}
\end{figure}

Figure~\ref{fig:difficulty} bins episodes by initial goal distance.
Mixed INT4 is higher than uniform INT4 in most bins, suggesting the 4-bit allocation effect is not confined to a single difficulty slice.

\begin{table}[H]
\centering
\small
\begin{tabular}{lcccc}
\toprule
Budget & Goal-distance bin & $n$ & Uniform INT4 & Mixed INT4 \\
\midrule
bA & low tertile & 10 & 0.0 & 0.3 \\
bA & mid tertile & 10 & 0.0 & 0.3 \\
bA & high tertile & 10 & 0.2 & 0.2 \\
bB & lower half & 10 & 0.3 & 0.4 \\
bB & upper half & 10 & 0.1 & 0.6 \\
\bottomrule
\end{tabular}
\caption{Difficulty-conditioned success from paired episodes. Directional mixed-vs-uniform gains appear across bins in both budgets.}
\label{tab:difficultybins}
\end{table}

\noindent\textbf{Inference.}
The mixed-over-uniform direction appears in low and medium bins and does not reverse in hard bins.
This suggests the allocation effect is not confined to one narrow difficulty regime, which improves confidence in transfer to broader evaluation sets.

\subsection{Mechanistic Context}
\begin{figure}[H]
  \centering
  \includegraphics[width=0.95\linewidth]{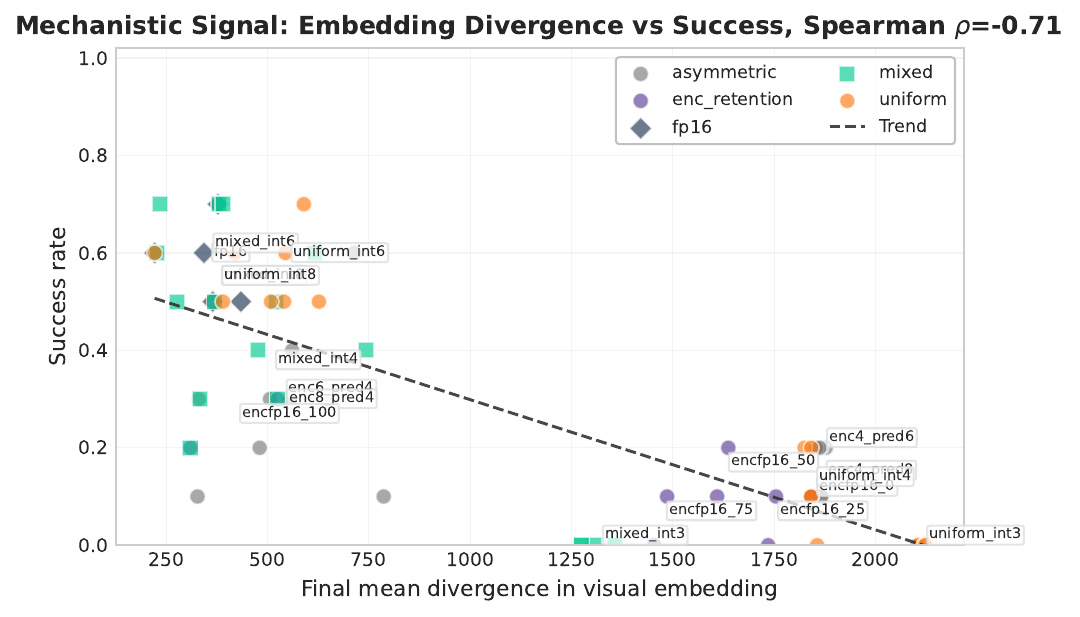}
  \caption{Mechanistic scatter: larger visual-embedding divergence is associated with lower planning success.}
  \label{fig:mech}
\end{figure}

Figure~\ref{fig:mech} links variant-level success to divergence metrics extracted from planning logs.
Correlations are computed over 65 run-level points (variant $\times$ budget $\times$ seed), with pooled budgets.
The negative trend supports a representation-degradation explanation for low-bit failure near the 4-bit boundary, while remaining correlational.

\noindent\textbf{Inference.}
The negative trend links quantization behavior to representation quality rather than only planner noise.
This does not establish causality, but it narrows follow-up experiments toward encoder-preserving or geometry-preserving quantization strategies.

\subsection{Mixed-Bit Maps for Hypothesis Generation}
\begin{figure}[H]
  \centering
  \includegraphics[width=0.95\linewidth]{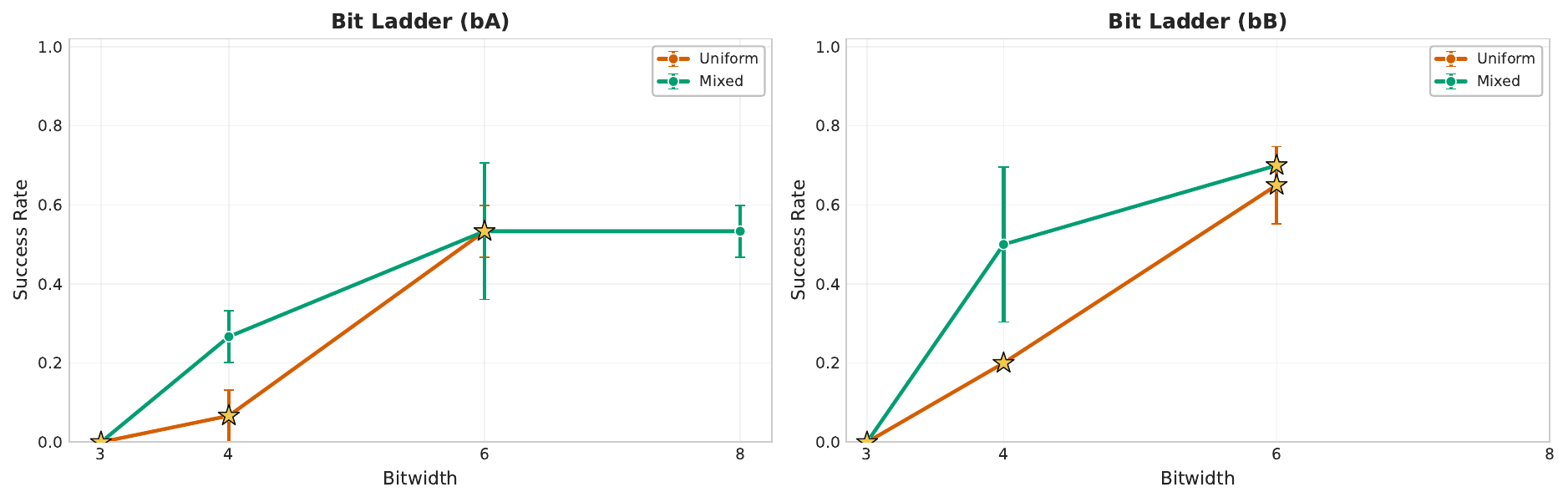}
  \caption{Uniform-vs-mixed bit ladder across budgets. Stars denote Pareto points in success--size space projected onto this ladder view.}
  \label{fig:ladder}
\end{figure}

\begin{figure}[H]
  \centering
  \includegraphics[width=0.95\linewidth]{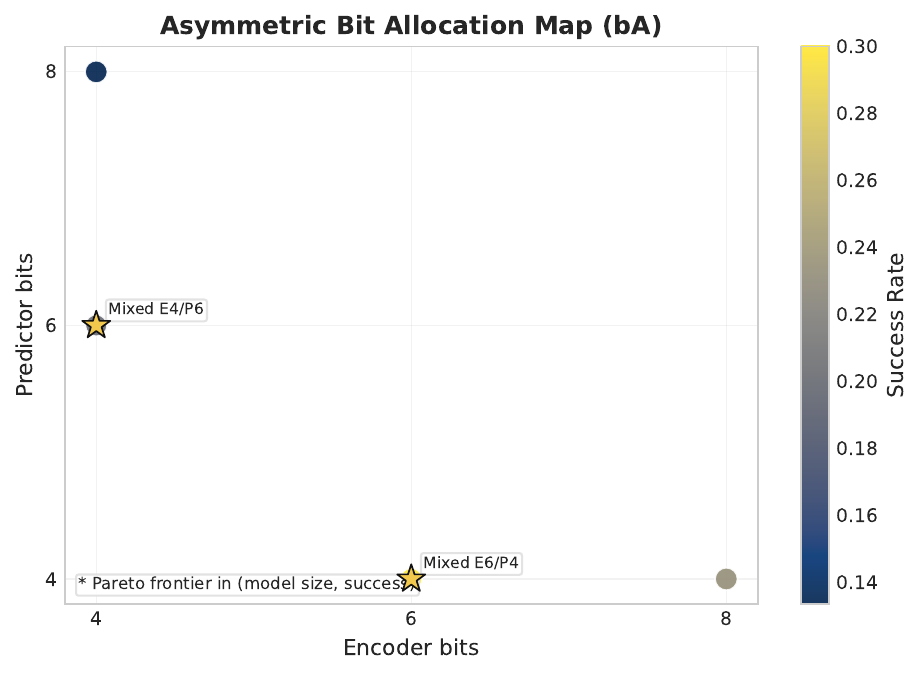}
  \caption{Asymmetric allocation map at bA. Marker color indicates success; star markers denote variants on the success--size Pareto frontier.}
  \label{fig:asym}
\end{figure}

The next two figures are exploratory and hypothesis-generating.
Figure~\ref{fig:ladder} visualizes the uniform/mixed bit ladder directly; Figure~\ref{fig:asym} maps asymmetric encoder/predictor allocations.
Together they motivate a budgeted bit-allocation optimization problem for future work.

\noindent\textbf{Inference.}
The ladder view highlights a phase-transition-like structure: stable high-bit behavior, a sharp low-bit collapse, and a sensitive middle regime.
The regime structure motivates adaptive bit-allocation policies instead of one-shot uniform compression.

\noindent\textbf{Inference.}
At comparable predictor precision, moving encoder precision upward typically improves outcomes more than the reverse.
Pareto-starred points concentrate in encoder-heavier configurations, reinforcing the encoder-priority conclusion from Tables~\ref{tab:paired} and \ref{tab:sizefair}.

\begin{figure}[H]
  \centering
  \includegraphics[width=0.95\linewidth]{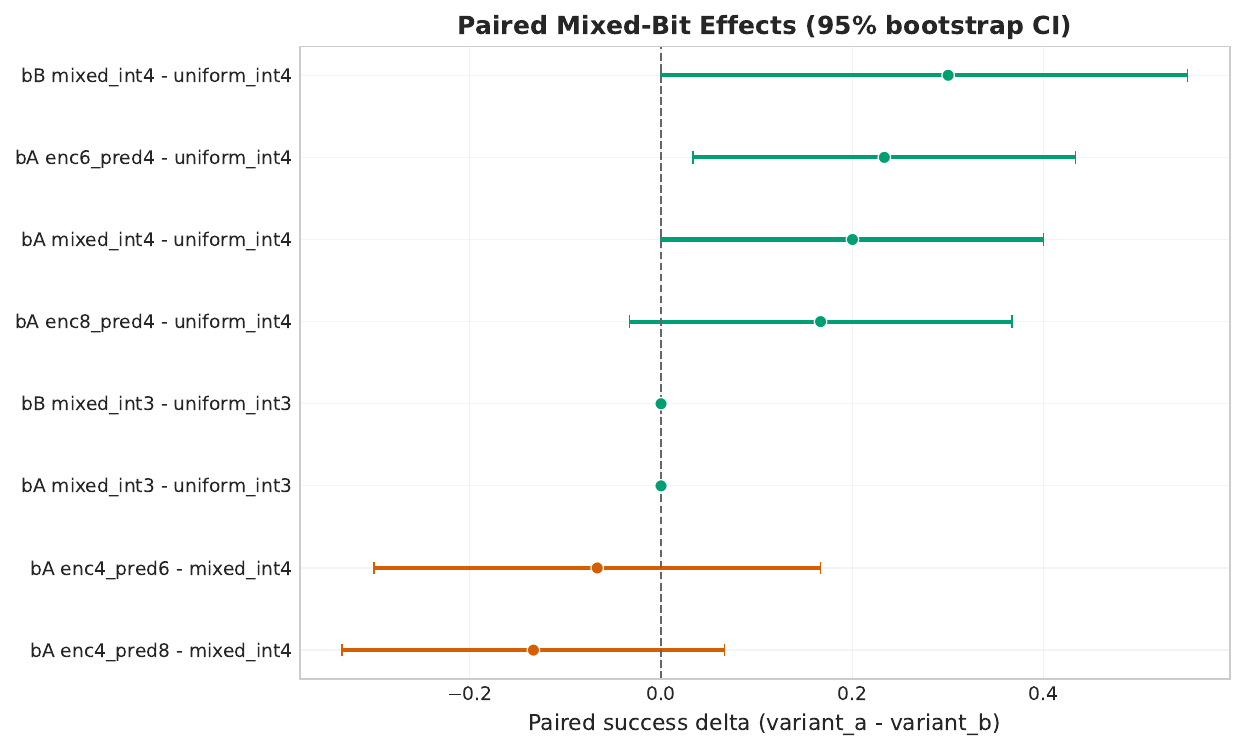}
  \caption{Forest plot of paired deltas with bootstrap confidence intervals for key comparisons.}
  \label{fig:forest}
\end{figure}

\noindent\textbf{Inference.}
The forest view makes effect direction and uncertainty transparent in one place.
Positive deltas for mixed-vs-uniform INT4 and encoder-heavy-vs-uniform comparisons support the framing that allocation matters most in the transition regime.

\end{document}